\def\year{2020}\relax
\documentclass[letterpaper]{article} 
\usepackage{aaai20}  
\usepackage{times}  
\usepackage{helvet} 
\usepackage{courier}  
\usepackage[hyphens]{url}  
\usepackage{graphicx} 
\usepackage{graphicx}  
\title{Order-free Learning Alleviating Exposure Bias in Multi-label Classification}
\author{Che-Ping Tsai and Hung-Yi Lee\thanks{This work was financially supported by the Ministry of Science and Technology of Taiwan.}\\  
Speech Processing and Machine Learning Laboratory , \\ National Taiwan University\\ 
r06922039@ntu.edu.tw, tlkagkb93901106@gmail.com }

\usepackage{amsfonts} 
\usepackage{amsmath}
\usepackage{multirow}
\usepackage{xcolor,colortbl}
\definecolor{Gray}{gray}{0.9}
\usepackage{comment}
\usepackage{afterpage}
\usepackage{svg}

\DeclareMathOperator{\sigmoid}{sigmoid}
\DeclareMathOperator{\softmax}{softmax}
\DeclareMathOperator{\LSTM}{LSTM}
\DeclareMathOperator{\BiLSTM}{BiLSTM}
\DeclareMathOperator{\MLP}{MLP}
\DeclareMathOperator{\ebF1}{ebF1}

\newcommand{\citet}[2][]{
   \citeauthor{#2} (\citeyear{#2})
}

\begin{document}

\maketitle

\begin{abstract}
Multi-label classification (MLC) assigns multiple labels to each sample. Prior studies show that MLC can be transformed to a sequence prediction problem with a recurrent neural network (RNN) decoder to model the label dependency. However, training a RNN decoder requires a predefined order of labels, which is not directly available in the MLC specification. Besides, RNN thus trained tends to overfit the label combinations in the training set and have difficulty generating unseen label sequences. 
In this paper, we propose a new framework for MLC which does not rely on a predefined label order and thus alleviates exposure bias. The experimental results on three multi-label classification benchmark datasets show that our method outperforms competitive baselines by a large margin.
We also find the proposed approach has a higher probability of generating label combinations not seen during training than the baseline models. 
The result shows that the proposed approach has better generalization capability.

\end{abstract}

\section{Introduction}
\label{sec:introduction}
Multi-label classification (MLC) is a fundamental but challenging problem in machine learning with applications such as text categorization~\cite{yang2018sgm}, sound event detection~\cite{gemmeke2017audio,yu2018multi}, and image classification~\cite{tsai2018adversarial}. 
In contrast to single-label classification, multi-label predictors must not only relate labels with the corresponding instances, but also exploit the underlying label structures. 
Take for instance the text classification dataset RCV1~\cite{lewis2004rcv1}, which uses a hierarchical tree structure between labels.

Recent studies show that MLC can be transformed to a sequence prediction problem by probabilistic classifier chains (PCC)\cite{read2011classifier,cheng2010bayes,dembczynski2012analysis}.
PCC models the joint probabilities of output labels with the use of the chain rule and predicts labels based on previously generated output labels.
Furthermore, PCC can be replaced by a RNN decoder to model label correlation.
\citet{wang2016cnn} propose the CNN-RNN architecture to capture both image-label relevance and semantic label dependency in multi-label image classification.
\citet{nam2017maximizing} and \citet{yang2018sgm} show that state-of-the-art multi-label text classification results can be achieved by using a sequence-to-sequence (seq2seq) architecture to encode input text sequences and decode labels sequentially. 

However, this kind of RNN-based decoder suffers from several problems. First, these models are trained using maximum likelihood estimation (MLE) on target label sequences, which relies on a predefined ordering of labels.
Previous studies~\cite{44871,yang2018deep} show that ordering has a significant impact on the performance. 
This issue also appears in the PCC,
It is addressed by ensemble averaging \cite{read2011classifier,cheng2010bayes}, ensemble pruning \cite{li2013selective}, pre-analysis of the label dependencies by Bayes nets  \cite{sucar2014multi} and integrating beam search with training to determine a suitable tag ordering \cite{kumar2013beam}. 
However, these approaches rely on training multiple models to ensemble or determine a proper order, which is computationally expensive.

Although \citet{nam2017maximizing} and \citet{yang2018sgm} compare several ordering strategies and suggest ordering positive labels by frequency directly in descending order (from frequent to rare labels), it is unnatural to impose a strict order on labels, which may break down label correlations in a chain. 
Furthermore, we find that this kind of model tends to overfit to label combinations and shows poor generalization ability.

Second, during training, the RNN-based models are always conditioned on correct prefixes; during inference, however, the prefixes are generated by the RNN-based model, yielding a problem known as \textit{exposure bias}~\cite{ranzato2015sequence} in seq2seq learning. The error may propagate as the model might be in a part of the state space that it has not seen during training~\cite{senge2014problem}.

In this paper, we propose a novel learning algorithm for RNN-based decoders on multi-label classification not rely on a predefined label order. 
The proposed approach is inspired by optimal completion distillation (OCD)~\cite{sabour2018optimal}, a training procedure for
optimizing seq2seq models. 
In this algorithm, we feed the RNN decoder generated tokens by sampling from the current model. Hence, the model may encounter different label orders and wrong prefixes during training, so the model explores more, and exposure bias is alleviated. 

Another common and straightforward way to avoid the need for ordered labels in MLC is binary relevance (BR)~\cite{tsoumakas2007multi}, which decomposes MLC into multiple independent single-label binary classification problems. 
However, this yields a model that cannot take advantage of label co-occurrences.
In this paper, we further propose helping the model to learn better by use of an auxiliary binary relevance (BR) decoder jointly trained with the RNN decoder within a multitask learning (MTL) framework. 

In addition, at the inference stage, the predictions of the BR decoder can be jointly combined in the RNN decoder to improve performance further. We propose two methods to combine their probabilities.
Extensive experiments show that the proposed model outperforms competitive baselines by a large margin on three multi-label  classification benchmark datasets, including two text classification and one sound event classification datasets.

The contributions of this paper are as follows:
\begin{itemize}

\item We propose a novel training algorithm for multi-label classification which predicts labels autoregressively but does not require a predefined label order for training.

\item We compare our methods with competitive baseline models on three multi-label classification datasets and demonstrate the effectiveness of the proposed models.

\item We systematically analyze the problem of exposure bias and the effectiveness of scheduled sampling~\cite{bengio2015scheduled} in multi-label classification.

\end{itemize}

\section{Related work}
\label{sec:related_work}
\subsection{RNN-based multi-label classification}
To free the RNN-based MLC classifier from a predefined label order, \citet{chen2017order} proposes the order-free RNN to dynamically decide a target label at each time during training by choosing the label in the target label set with the highest predicted probability; hence, the model learns a label order by itself. 
Although the order can be modified during training, this approach still needs an initialized label order to start the training process. 
We find order-free RNN shows poor generalization ability to unseen label combinations in the experiments.
Also, as the model is always supplied with the correct labels, it suffers from exposure bias.

To handle both exposure bias and label order, other studies apply a reinforcement learning (RL) algorithm to MLC. 
\citet{he2018reinforced} apply an off-policy Q learning algorithm to multi-label image classification.  \citet{yang2018deep} uses two decoders to solve multi-label text classification, one of which is trained with MLE and the other is trained with a self-critical policy gradient training algorithm. However, Q learning and policy gradients cannot easily incorporate ground truth sequence information, except via the reward function, as the model is rewarded only at the end of each episode. 
Indeed, \citet{he2018reinforced} does not work without pretraining on the target dataset. 
By contrast, we use optimal completion distillation (OCD)~\cite{sabour2018optimal} for MLC, which optimizes token-level log-loss, where the training is stabilized and requires neither initialization nor joint optimization with MLE.

\subsection{Optimal Completion Distillation (OCD)}
Our work is inspired by OCD~\cite{sabour2018optimal}, which was first used in the context of end-to-end speech recognition in which it achieved state-of-the-art performance. 
In contrast to MLE, OCD algorithms encourage the model to extend all possible tokens that lead to the optimal edit distance by assigning equal probabilities to the target policy that the model learns from. 
We use OCD to train the RNN decoder in MLC. 
The OCD training details for MLC are in section \emph{Learning for RNN decoder $\mathcal{D}_{rnn}$}.
In contrast to the original OCD~\cite{sabour2018optimal} which optimizes the edit distance, in the proposed approach we optimize the numbers of missing and false alarm labels.

\section{Model architecture}
\label{sec:model_architecture}
\begin{figure}
  \includegraphics[width=\linewidth]{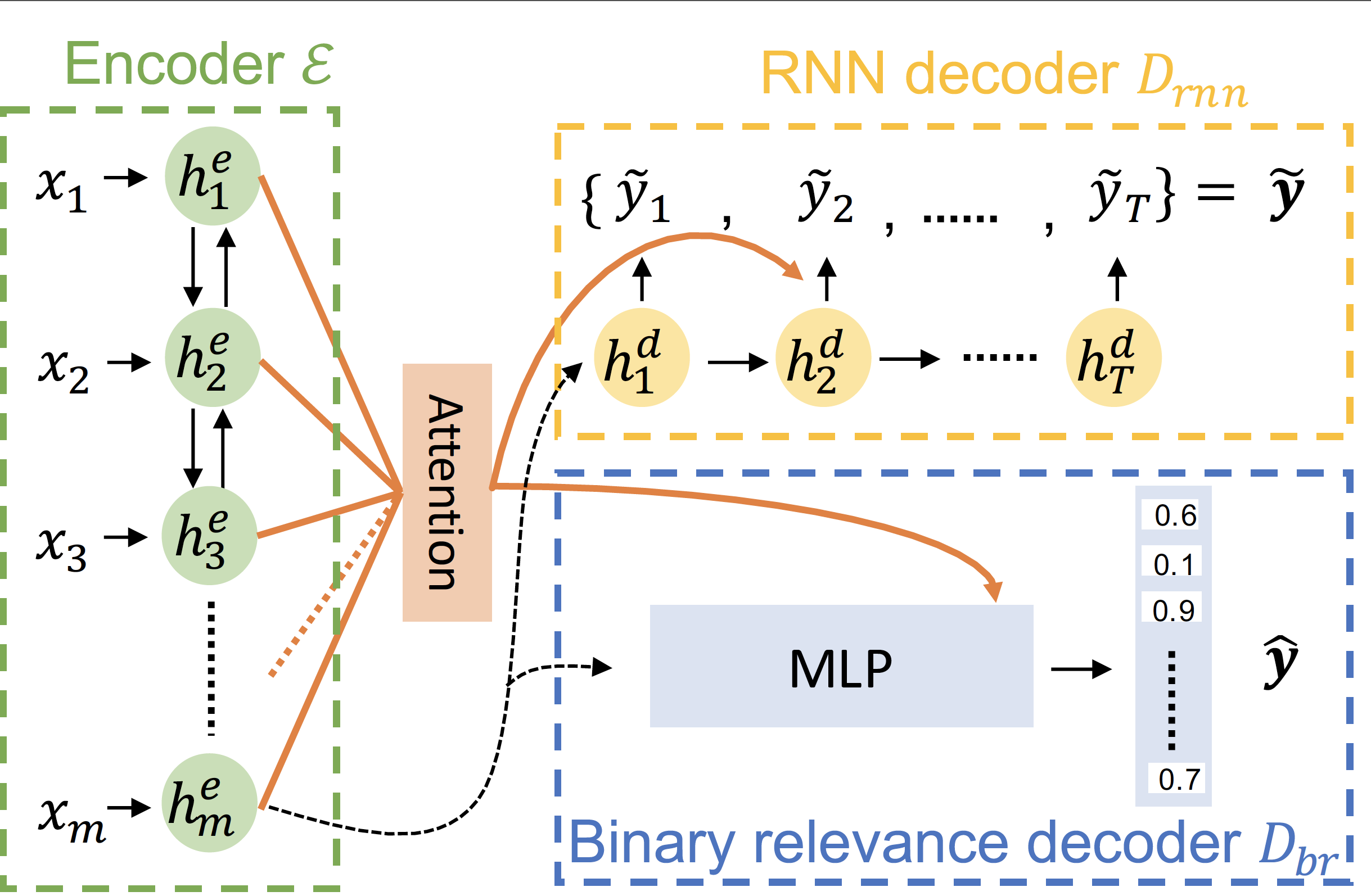}
  \caption{Overview of proposed model. The model is composed of three components: encoder $\mathcal{E}$, RNN decoder $\mathcal{D}_{rnn}$, and binary relevance decoder $\mathcal{D}_{br}$. $\tilde{\mathbf{y}}=\{\tilde{y}_1, \tilde{y}_2,...,\tilde{y}_{T} \}$ represents the sampled sequence from $\mathcal{D}_{\mathit{rnn}}$, while $\hat{y}$ is a vector representing the predicted probabilities of labels by $\mathcal{D}_{br}$.}
  \label{fig:model}
\end{figure}

An overview of the proposed model is shown in Fig.~\ref{fig:model}. The model is composed of three components: encoder $\mathcal{E}$, RNN decoder $\mathcal{D}_{rnn}$, and binary relevance decoder $\mathcal{D}_{br}$. Here, multi-label text classification is considered an instance of MLC. For other types of MLC other than text classification (e.g. sound event classification), the architecture of the encoder can be changed.

\subsection{Encoder  $\mathcal{E}$}
We employ a bidirectional LSTM as an encoder $\mathcal{E}$. The encoder reads the input text sequence $\mathbf{x} = {x_1,x_2,...,x_m}$ of $m$ words in both forward and backward directions and computes the hidden states $h_{1}^e,h_{2}^e,...h_{m}^e $  for each word.
\begin{equation}
h_{1}^e,h_{2}^e,...h_{m}^e = \BiLSTM(x_1,x_2,...,x_m)
\label{eqn:encoder_lstm}
\end{equation}

\subsection{RNN decoder $\mathcal{D}_{rnn}$}
\label{sec:rnn_decoder}
The RNN decoder seeks to predict labels sequentially. It is potentially more powerful than the binary relevance decoder because each prediction is determined based on the previous prediction: thus it implicitly learns label dependencies. We implement it using LSTMs with an attention mechanism. Hence, the encoder and RNN decoder form a seq2seq model.
In particular, we set the initial hidden state of the decoder $h_{0}^d = h_{m}^e$ and calculate the hidden state $h_{t}^d$ and output $o_t$ at time $t$ as 

\begin{equation}
h_{t}^d, o_t = \LSTM(h^d_{t-1},\tilde{y}_{t-1})
\label{eqn:rnn_decoder}
\end{equation}
 where $\tilde{y}_{t-1}$ is the predicting label at previous timestep.
 $\tilde{y}_t$ is estimated with $p_{rnn}(\tilde{y}_t|\tilde{\mathbf{y}}_{<t},\mathbf{x})$ by the following equations:
\begin{equation}
p_{rnn}(\tilde{y}_t|\tilde{\mathbf{y}}_{<t},\mathbf{x}) = \softmax(o_t)
\label{eqn:rnn_decoder_prnn}
\end{equation}
\begin{equation}
\tilde{y}_t \sim \softmax(o_t+M_t),
\label{eqn:rnn_decoder_mask}
\end{equation}
During sampling, we add a mask vector $M_t \in \mathcal{R}^L$ \cite{yang2018sgm}  to prevent the model from predicting repeated labels, where $L$ is the number of labels in the dataset:
\begin{equation}
(M_t)_j=
\begin{cases}
-\infty & \text{if the $j$-th label has been predicted}\\
& \text{before step $t$.}\\
0 & \text{otherwise.}
\end{cases}
\label{rnn_decoder8}
\end{equation}

\subsection{Binary relevance decoder $\mathcal{D}_{br}$}
The binary relevance (BR) decoder $\mathcal{D}_{br}$  here is an auxiliary decoder to train the encoder within the multitask learning (MTL) framework. The BR decoder predicts each label separately as a binary classifier for each label, helping the model to learn better.
Another advantage of using the BR decoder is that we can consider the predictions of both the RNN and BR decoders to further improve performance.

In particular, we feed the final hidden state of encoder $h_{m}^e$ to a DNN with a final prediction layer of size $L$ with sigmoid activation functions to predict the probabilities of each label. 
To take into account vanishing gradients for long input sequences, we add another attention module. In particular, we calculated the context vector $c^{br}$ in the attention mechanism with the output of fully-connected layers $\MLP(h_m^{e})$ and then compute probabilities  $p_{br}(\tilde{y}|\mathbf{x})$ as

\begin{equation}
p_{br}(\tilde{y}|\mathbf{x}) = \sigmoid(W_{br}[\MLP(h_m^{e});c^{br}]),
\label{eqn:br_decoder}
\end{equation}
where $W_{br}$ is the 
matrix of weight parameters and $[\MLP(h_m^{e});c^{br}]$ indicates the concatenation of $\MLP(h_m^{e})$ and $c^{br}$.

\section{Order-Free Training} 
\label{sec:training}
In this section, we derive the training objective for the RNN decoder  , the BR decoder , and the multitask learning objective.

\subsection{Learning for RNN decoder $\mathcal{D}_{rnn}$}
\label{sec:loss_rnn_decoder}

\subsubsection{RNN decoder learning as RL}

To reduce exposure bias and free the model from relying on a predefined label order, we never train on ground truth target sequences. Instead, we approach the MLC problem from an RL perspective. 
The model here plays the role of an \textit{agent} whose action $a_{t}$ is the current generated label at time $t$ and whose \textit{state} $s_{t}$ is the output labels $\tilde{\mathbf{y}}_{<t}$ before time $t$. The policy $\pi(s)$ is  a probability distribution over actions $a$ given states $s$. Once the process is ended with an \textit{end-of-sentence} token, the agent is given a \textit{reward} $R$. 

In our approach, reward $R$ is defined as 
\begin{equation}
R(\mathbf{y}^{*},\tilde{\mathbf{y}}) = - |\{ \mathbf{y}^{*}\} \setminus \{ \tilde{\mathbf{y}}\}| - | \{ \tilde{\mathbf{y}}\}\setminus \{ \mathbf{y}^{*}\}|,
\label{eq:reward}
\end{equation}
where $\mathbf{y}^{*}$ and $\tilde{\mathbf{y}}$ are the ground truth labels and the sequence of labels generated by the RNN decoder, and $B \setminus A$ is the relative complement of set A in set B. 

The first and second term of reward $R(\mathbf{y}^{*},\tilde{\mathbf{y}})$ are the number of labels that were not predicted and the number of misclassified labels, respectively.

\subsubsection{Optimal completion distillation}
\label{sec:OCD}

However, typical RL algorithms, such as Q learning and policy gradients, cannot easily incorporate ground truth sequence information except via the reward function.
Here we introduce optimal Q-values, which evaluates each action $a_t$ at each time $t$.

Optimal Q-values $Q^{*}(s,a)$ represents the maximum total reward the agent can acquire after taking action $a$ at state $s$ via subsequently conducting the optimal action sequence. Optimal Q-values at time $t$ can be expressed as  
\begin{equation}
	 Q^{*}(\tilde{\mathbf{y}}_{<t},a) = \max_{\mathbf{y}_{opt} \in \mathcal{Y}} R(\mathbf{y}^{*},[\tilde{\mathbf{y}}_{<t}, a, \mathbf{y}_{opt}]).
\label{eq:optimal_q_value}
\end{equation}
where $[\tilde{\mathbf{y}}_{<t}, a, \mathbf{y}_{opt}]$ is a complete sequence, which is the concatenation of token sequence generated before time $t$, action at time $t$  and optimal subsequent action sequence $\mathbf{y}_{opt}$.

Optimal policy at time $t$ can be calculated by taking a softmax over optimal Q-values of all the possible actions. Formally, 
\begin{equation}
	\pi^{*}(a|\tilde{\mathbf{y}}_{<t})=\frac{\exp({Q^{*}(\tilde{\mathbf{y}}_{<t},a)/\tau})} {\sum_{a'} \exp(Q^{*}(\tilde{\mathbf{y}}_{<t},a')/\tau)},
\label{eq:optimal_policy}
\end{equation}
where $\tau \geq 0 $  is a temperature parameter. If $\tau$ is close to $0$, $\pi^{*}$ is a hard target.
Table~\ref{tab:ocd_example} shows an example illustrating the optimal policy in OCD training procedure.


Given a dataset $\{\mathbf{x},\mathbf{y}^{*}\}$, we first draw generated sequences from the RNN decoder $\tilde{\mathbf{y}} \sim p_{rnn}(\cdot|\mathbf{x})$ by sampling.
The loss function $\mathcal{L}_{\mathit{OCD}}$ can be obtained via calculating KL divergence between the optimal policy and the model's generated distribution over labels at every time step $t$, 
\begin{equation}
    \begin{split}
    & \mathcal{L}_{\mathit{OCD}} = \mathbb{E}_{(\mathbf{x},\mathbf{y}^{*})\sim data} \mathbb{E}_{\tilde{\mathbf{y}} \sim p_{rnn}(\cdot|\mathbf{x}) } [\\
    & \sum_{t=1}^{|\tilde{\mathbf{y}}|}\mathrm{KL}( \pi^{*}(\cdot|\tilde{\mathbf{y}}_{<t})  || p_{rnn}(\cdot | \tilde{\mathbf{y}}_{<t},\mathbf{x}))].
    \end{split}
\label{eq:ocd}
\end{equation}
The above equation means we ``distill'' knowledge from optimal policies obtained by completing with optimal action sequences to RNN decoder $p_{rnn}$, so RNN decoder can have similar behaviour as optimal policy $\pi^{*}$.  

In contrast to MLE, OCD encourages the model to extend all the possible targets resulting in the same evaluation metric score. Therefore, the OCD objective focuses on all labels that were not predicted and assigns them equal probabilities. Once all the target labels are successfully generated, the objective guides the model to produce the end-of-sentence token with probability $1$.  

Since the OCD targets depend only on the tokens generated previously,  we do not need a human-defined label order to train the RNN decoder. The label order is instead automatically
determined at each time step. In addition, we always train on sequences generated from the current model, thus alleviating exposure bias. Note that we can substitute the reward function (Eq.~\ref{eq:reward}) with other example-based test metrics such as the example-based F1 score (Eq.~\ref{eq:ebf1} in the Appendix), but these lead to the same OCD targets as the rewards of all the target labels are the same. 

\begin{table}[h!]
\renewcommand{\arraystretch}{1.15}
\begin{tabular}{|c|c|c|c|}
 \hline
   \cellcolor[gray]{0.9}{Time} & \cellcolor[gray]{0.9}{OCD} & \cellcolor[gray]{0.9}{Optimal policy} & \cellcolor[gray]{0.9}{Prediction} \\ \cellcolor[gray]{0.9}{$t$} & \cellcolor[gray]{0.9}{targets}  & \cellcolor[gray]{0.9}{$\pi^{*}$ ($\tau \to 0$)} &
   \cellcolor[gray]{0.9}{$\tilde{y}_t$} \\ 
 \hline
 0 & A, B, D & $[\frac{1}{3},\frac{1}{3},0,\frac{1}{3},0]$ & B \\
 \hline
 1 & A, D & $[\frac{1}{2},0,0,\frac{1}{2},0]$ & C \\
 \hline
 2 & A, D & $[\frac{1}{2},0,0,\frac{1}{2},0]$ & A \\
 \hline
 3 & D & $[0,0,0,1,0]$ & D \\
 \hline
 4 & \textless eos\textgreater & $[0,0,0,0,1]$ &\textless eos\textgreater \\
 \hline
\end{tabular}
\caption{A training example of optimal completion distillation, where there are 4 kinds of labels (A, B, C, D) and an end-of-sentence token (\textless eos\textgreater). Labels A, B, and D are the targets of this instance and the vectors of $\pi^{*}$ represent probabilities for labels A, B, C, D and \textless eos\textgreater  , respectively. We set $\tau \to 0 $ here, so the optimal policy only encourages labels with the highest optimal Q values. For example, at time 1, since the model has predicted correct token B at time 0, there are two optimal extended tokens \{A, D\}, which result in a total reward of 0  (Eq.~\ref{eq:reward}) when combined with proper suffixes.
Then we sample from the current policy and predict the incorrect token C, which leads to a decreased optimal possible reward of $-1$ (Eq.~\ref{eq:reward}) at time 2.}
\label{tab:ocd_example}
\end{table}

\subsection{Learning for binary relevance decoder $\mathcal{D}_{br}$}
\label{sec:loss_br}

For the binary relevance decoder, given the ground truth $\mathbf{y}^{*}_{vec} = [ y^{*}_{vec,1},y^{*}_{vec,2}...,y^{*}_{vec,L}]^T \in \{0,1\}^{L}$ in binary format, we use binary cross-entropy loss as the objective:
\begin{equation}
	\begin{split}
 		& \mathcal{L}_{\mathit{logistic}} =\mathbb{E}_{( \mathbf{x},\mathbf{y}^{*}_{vec})\sim{\mathit{data}}} [  \sum_{i=1}^{L} y^{*}_{vec,i} log\,\hat{y}_{i} +\\
        & (1-y^{*}_{vec,i})log(1-\hat{y}_{i}) ],
	\end{split}
\label{eq:logistic_loss} 
\end{equation}
where $\hat{\mathbf{y}} = [ \hat{y}_{1},\hat{y}_{2}...,\hat{y}_{L}]^{T}$, which is a vector of length $L$  representing the predicted probability of each label. 

\subsection{Multitask Learning (MTL)}
\label{sec:mtl_loss}
The objective of MTL is
\begin{equation}
    \mathcal{L}_{\mathit{MTL}} = \mathcal{L}_{\mathit{OCD}} +     \lambda\mathcal{L}_{\mathit{logistic}},
\end{equation}
where $\lambda$ is a weight.

\section{Decoder Integration} 
\label{sec:testing}
In this section, we seek to utilize both the RNN and BR decoders to find the optimal hypothesis $\mathcal{H} =\{l_1,l_2,...,l_T\}$, which consists of $T-1$ predicted labels $\{l_1,l_2,...,l_{T-1}\}$, where $l_T$ is the end-of-sentence token indicating that the decoding process of the RNN decoder has ended.

For the BR decoder, the outputs after a sigmoid activation $ p_{br}(y_l = 1|\mathbf{x}) = \hat{y}_{l} $ are designed to estimate the posterior probabilities of each label $l$. Therefore, the theoretically optimal threshold for converting the probability to a binary value should be 0.5, which is equivalent to finding the optimal hypothesis $\mathcal{H}$ that maximizes the Eq.~\ref{eq:p_br} below, which is the product of the probabilities of all the labels.
\begin{equation}
P_{br} (\mathcal{H})= \prod_{l \in \mathcal{H}}p_{br}(y_{l}=1|\mathbf{x}) \times \prod_{l \notin \mathcal{H}}p_{br}(y_{l}=0|\mathbf{x})
\label{eq:p_br}
\end{equation}


For the RNN decoder, a typical inference step is performed with a beam search to solve Eq.~\ref{eq:p_path}~\cite{yang2018sgm,chen2017order}. Given input $x$, the probability of the predicted hypothesis path $\mathcal{H}$ is
\begin{equation}
P_{path}(\mathcal{H}) = \prod_{i=1}^{i=T}p_{rnn}(l_i|\mathbf{x},l_1,...l_{i-1}).
\label{eq:p_path}
\end{equation}

To combine the predictions of the RNN and BR decoders,  we simply take the product of Eq.~\ref{eq:p_br} and Eq.~\ref{eq:p_path} to yield the final objective function Eq.~\ref{eq:p_joint}: 

\begin{equation}
\hat{\mathcal{H}} = \arg \max_{\mathcal{H}} \{ P_{path}(\mathcal{H}) \times P_{br}(\mathcal{H}) \}.
\label{eq:p_joint}
\end{equation}

Nonetheless, the equation is not easy to solve because the RNN decoder outputs the probability of selecting a particular label at each time step while the BR decoder produces the unconditional probabilities of all the labels at once. To incorporate the BR probabilities of labels in the score, we provide two different decoding strategies to find the best hypothesis $\hat{\mathcal{H}}$ .

\subsection{Logistic Rescoring}

In this method, we first obtain a set of complete hypotheses using beam search only with the RNN decoder, and rescore each hypothesis $\mathcal{H}$ using the probabilities produced by the BR decoder with Eq.~\ref{eq:p_br}. Finally, we select as the final prediction the hypothesis $\mathcal{H}$ with the highest $P_{br}(\mathcal{H}_{best})$.

\subsection{Logistic Joint Decoding}
This method is one-pass decoding. We conduct a beam search according to the following equation (see the derivation in Appendix~\ref{sec:derivation}).

\begin{equation}
P_{joint}(\mathcal{H}) = \prod_{i=1}^{i=T}p_{rnn}(l_i|\mathbf{x},l_1,...l_{i-1})\frac{p_{br}(y_{l_i}=1|\mathbf{x})}{p_{br}(y_{l_i}=0|\mathbf{x})}  
\label{eq:br_joint_decoding}
\end{equation}

Note that we manually set the probability of the end-of-sentence token $l_T$ for binary relevance $p_{br}(y_{l_T} = 0|\mathbf{x}) = p_{br}(y_{l_T} = 1|\mathbf{x}) = 0.5$, since it does not exist in the outputs of the BR decoder.

\section{Experimental Setup}
\label{sec:experimental_setup}
We validate our proposed model on two multi-label text classification datasets, which are AAPD~\cite{yang2018sgm} and Reuters-21758, and a sound event classification dataset, which is Audio set \cite{gemmeke2017audio} proposed by Google. They span a wide variety in terms of the number of samples, the number of labels, and the number of words per sample. 
Due to space limit, we put an extra experiment in text classification on RCV1-V2 \cite{lewis2004rcv1}, data statistics, experimental settings  and the descriptions of five evaluation metrics in Appendix.

\begin{table}
\centering
\resizebox{1.08\columnwidth}{!}{
\begin{tabular}{|l|l|c|c|c|c|c|c|}
\hline
\rowcolor{Gray}
\multicolumn{2}{|c|}{Models} & maF1 & miF1 & ebF1 & ACC & HA & Average \\
\hline
\multicolumn{2}{|l|}{(a) Seq2set (simp.)}  &&& &&& \\
\multicolumn{2}{|l|}{\cite{yang2018deep}} & \multirow{-2}{*}{-} & \multirow{-2}{*}{0.705} & \multirow{-2}{*}{-} &  \multirow{-2}{*}{-} & \multirow{-2}{*}{0.9753} & \multirow{-2}{*}{-}  \\
\hline
\multicolumn{2}{|l|}{(b) Seq2set}  &&& &&& \\
\multicolumn{2}{|l|}{\cite{yang2018deep}} & \multirow{-2}{*}{-} & \multirow{-2}{*}{0.698} & \multirow{-2}{*}{-} &  \multirow{-2}{*}{-} & \multirow{-2}{*}{0.9751} & \multirow{-2}{*}{-}  \\
\hline
\multicolumn{2}{|l|}{(c) SGM+GE}  &&&&&&  \\
\multicolumn{2}{|l|}{\cite{yang2018sgm}} &
\multirow{-2}{*}{-} & \multirow{-2}{*}{0.710} & \multirow{-2}{*}{-} &  \multirow{-2}{*}{-} & \multirow{-2}{*}{0.9755} & \multirow{-2}{*}{-}  \\
\hline
\rowcolor{Gray}
\multicolumn{8}{|c|}{Baselines} \\
\hline
\multicolumn{2}{|l|}{(d) BR} & 0.523 & 0.694 & 0.695 & 0.368	& 0.9741 & 0.651 \\
\hline
\multicolumn{2}{|l|}{(e) BR++} & 0.521 & 0.700 & 0.703 & 0.390 & 0.9750 & 0.658 \\
\hline
\multicolumn{2}{|l|}{(f) Seq2seq} & 0.511 & 0.695	 & 0.707 & \textbf{0.421} & 0.9743 &	0.662 \\
\hline
\multicolumn{2}{|l|}{(g) Seq2seq + SS} & 0.541 & 0.703 &	0.713 & 0.406 & 0.9742 & 0.667 \\
\hline
\multicolumn{2}{|l|}{(h) Order-free RNN} & 0.539 & 0.696 & 0.708 & 0.413 & 0.9742 & 0.666 \\
\hline
\multicolumn{2}{|l|}{(i) Order-free  RNN +  SS} & 0.548	& 0.699	& 0.709 &	0.416 & 0.9743 & 0.669	\\
\hline
\rowcolor{Gray}
\multicolumn{8}{|c|}{Proposed methods} \\
\hline
\multicolumn{2}{|l|}{(j) OCD} & 0.541 & \textcolor{blue}{0.707} & \textcolor{blue}{0.723} & 0.403 & 0.9740 & \textcolor{blue}{0.670} \\
\hline
\multirow{4}{*}{\shortstack{OCD\\+\\MTL}} & (k) RNN dec. & \textcolor{blue}{0.578} & \textcolor{blue}{0.711} & \textcolor{blue}{0.727} & 
0.391 & 0.9742 & \textcolor{blue}{0.676} \\
\cline{2-8}
&(l) BR dec. & \textcolor{blue}{0.562} & 0.711 &	\textcolor{blue}{0.718} & 0.382 & \textcolor{blue}{\textbf{0.9760}} & \textcolor{blue}{0.670} \\
\cline{2-8}
&(m) Logistic rescore & \textcolor{blue}{\textbf{0.585}} & \textcolor{blue}{\textbf{0.720}} & \textcolor{blue}{\textbf{0.736}} & 0.395 & 0.9749 & \textcolor{blue}{\textbf{0.682}} \\
\cline{2-8}
&(n) Logistic joint dec. & \textcolor{blue}{0.580} & \textcolor{blue}{0.719} & \textcolor{blue}{0.731} & 0.399 & \textcolor{blue}{0.9753} & \textcolor{blue}{0.681} \\
\hline
\rowcolor{Gray}
\hline
\end{tabular}}
\caption{Performance on AAPD}
\vspace{-0.2cm}
\label{tab:aapd_performance}
\end{table}

\subsection{Evaluation Metrics}
Multi-label classification can be evaluated with multiple metrics, which capture different aspects of the problem. We follow~\citet{nam2017maximizing} in using five different metrics: subset accuracy (ACC), Hamming accuracy (HA), example-based F1 (ebF1), macro-averaged F1 (maF1), and micro-averaged F1 (miF1). 

\subsection{Baselines}

We compare our methods with the following baselines. For a fair comparison, the architecture of all the encoders are the same except for BR++: the RNN decoders for Seq2seq, Order-free RNN, and the proposed approaches are the same. 

\begin{itemize}
\item \textbf{Binary Relevance (BR)} is the model trained with logistic loss (Eq.~\ref{eq:logistic_loss}), and consists of an encoder and a BR decoder. 

\item \textbf{Binary Relevance++ (BR++)} is a model with a larger encoder but with the same training algorithm as BR. 
Because the MTL model has more parameters than BR, for fair comparison, we increase the number of layers in the encoder RNN so that the number of parameters is approximately equal to the MTL model.
\footnote{Since \citet{yu2018multi} have tested different architectures of BR models on Audio set, we do not use BR++ as a baseline.}

\item \textbf{Seq2seq}~\cite{nam2017maximizing} is composed of an RNN encoder and an RNN decoder with an attention mechanism. The model is trained with MLE. The target label sequences are ordered from frequent to rare, which yields better performance~\cite{nam2017maximizing,yang2018sgm}.
\footnote{For Audio set, the architecture of the encoder is described in Appendix and is not based on RNN. For consistency, we denoted it as Seq2seq.}

\item \textbf{Order-free RNN}~\cite{chen2017order} uses an algorithm for multi-label image classification to train the RNN decoder without predefined orders but suffers from exposure bias.
\end{itemize}

Since scheduled sampling also tackles the problem of exposure bias, we also compare the performance of seq2seq and order-free RNN with scheduled sampling (SS), which are \textbf{Seq2seq + SS} and \textbf{Order-free RNN + SS}. The detailed discussion of the effectiveness of scheduled sampling is in section \emph{Disussion.}

\section{Results and Discussion}
\label{sec:experimental_results}
\begin{figure}
  \includegraphics[width=\linewidth]{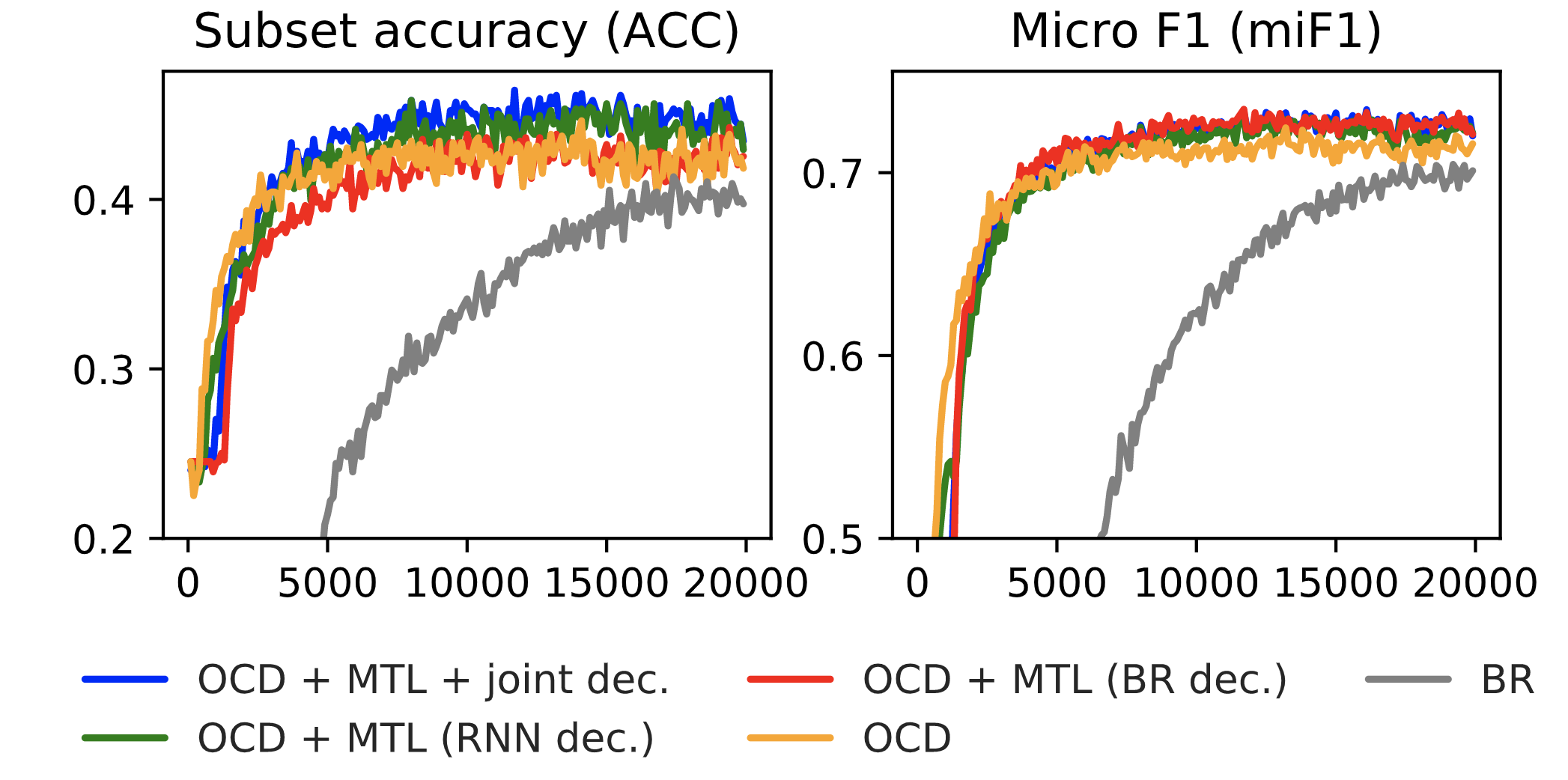}
  \caption{Performance of BR, OCD and MTL models on AAPD validation set with different decoding strategies during training. The x-axis denotes the number of updates; we use different scales on the y-axis for each measure.}
  \label{fig:training_curve}
\vspace{-0.2cm}
\end{figure}

\begin{table}
\centering
\resizebox{1.06\columnwidth}{!}{
\begin{tabular}{|c|c|c|c|c|c|c|c|}
\hline
\rowcolor{Gray}
\multicolumn{2}{|c|}{Models} & maF1 & miF1 & ebF1 & ACC & HA & Average \\
\hline

\multicolumn{2}{|c|}{SVM}  &&& &&& \\
\multicolumn{2}{|c|}{\cite{debole2005analysis}}  & \multirow{-2}{*}{0.468} & \multirow{-2}{*}{0.787} & \multirow{-2}{*}{-} &  \multirow{-2}{*}{-} & \multirow{-2}{*}{-} & \multirow{-2}{*}{-}  \\
\hline
\multicolumn{2}{|c|}{EncDec}  &&& &&& \\
\multicolumn{2}{|c|}{\cite{nam2017maximizing}}  & \multirow{-2}{*}{0.457} & \multirow{-2}{*}{0.855} & \multirow{-2}{*}{0.891} &  \multirow{-2}{*}{0.828} & \multirow{-2}{*}{0.996} & \multirow{-2}{*}{0.805}  \\
\hline
\rowcolor{Gray}
\multicolumn{8}{|c|}{Baselines} \\
\hline
\multicolumn{2}{|c|}{BR} & 0.442 & 0.861 & 0.878 & 0.817 & 0.9964 & 0.799 \\
\hline
\multicolumn{2}{|c|}{BR++} & 0.407 & 0.852 & 0.861 & 0.812 & 0.9962 & 0.786 \\
\hline
\multicolumn{2}{|c|}{Seq2seq} & 0.465 & 0.862	 & 0.895 & 0.834 & 0.9965 & 0.811 \\
\hline
\multicolumn{2}{|c|}{Seq2seq+SS} & 0.464 & 0.856 & 0.895 & 0.834 & 0.9965 & 0.809 \\
\hline
\multicolumn{2}{|c|}{Order-free RNN} & 0.445 & 0.862 & 0.901 & 0.835 & 0.9963 & 0.806 \\
\hline
\multicolumn{2}{|c|}{Order-free  RNN +  SS}	 & 0.452 & 0.859 & 0.896 & 0.836 & 0.9962 & 0.808 \\
\hline
\rowcolor{Gray}
\multicolumn{8}{|c|}{Proposed methods} \\
\hline
\multicolumn{2}{|c|}{OCD}& 0.458 & \textcolor{blue}{0.872} & 
\textcolor{blue}{0.903} & 
\textcolor{blue}{0.839} & 
\textcolor{blue}{0.9966} & 
\textcolor{blue}{0.814} \\
\hline
\multirow{4}{*}{\shortstack{OCD\\+\\MTL}} & RNN dec. & \textcolor{blue}{0.475} & 
\textcolor{blue}{0.874} & 
\textcolor{blue}{\textbf{0.905}} & 
\textcolor{blue}{\textbf{0.844}} & 
\textcolor{blue}{0.9966} & 
\textcolor{blue}{0.819} \\
\cline{2-8}
& BR dec. & 0.459	& 
\textcolor{blue}{\textbf{0.877} }& 0.898 & 0.835 & \textcolor{blue}{0.9966} & \textcolor{blue}{0.813} \\
\cline{2-8}
& Logistic rescore & 
\textcolor{blue}{0.477} & 
\textcolor{blue}{0.875} & 
\textcolor{blue}{0.903} &	
\textcolor{blue}{0.842} & 
\textcolor{blue}{\textbf{0.9967}} & 
\textcolor{blue}{0.819} \\
\cline{2-8}
& Logistic joint dec. & 
\textcolor{blue}{\textbf{0.490}} & 
\textcolor{blue}{0.874} & 
\textcolor{blue}{0.904} & 
\textcolor{blue}{0.843} & 
\textcolor{blue}{\textbf{0.9967}} & 
\textcolor{blue}{\textbf{0.822}} \\
\hline
\rowcolor{Gray}
\hline
\end{tabular}}
\caption{Performance comparisons on Reuters-21578}
\label{tab:reuters_performance}
\end{table}

In the following, we show results of the baseline models and the proposed method on three text datasets. For MTL models, we show the results of the four kinds of different decoding strategies described in section \emph{Decoder Integration}. For a simple comparison, we also compute averages of the five metrics as a reference. Note that blue an bold texts in Table \ref{tab:aapd_performance}-\ref{tab:audio_set} mean that the performance of proposed methods surpass all the baselines and the highest performance in each measure.

\subsection{Experiments on AAPD}

The experimental results on the AAPD dataset are shown in Table~\ref{tab:aapd_performance}. We see that different models are skilled at different metrics. For example, RNN decoder based models, i.e. Seq2seq in row (f) and Order-free RNN in row (h), perform well on ACC, whereas BR and BR++  have better results in terms of HA but show clear weaknesses in predicting rare labels (cf. especially maF1). However, OCD in row (j) performs better than all the baselines (row (d)--(i)) (0.672 on average),\footnote{Except for ACC: the reason is given in the following discussion.} especially in miF1 (0.707) and ebF1 (0.737), which verifies the power of the proposed training algorithm. 

\begin{table}
\centering
\resizebox{1.06\columnwidth}{!}{
\begin{tabular}{|c|c|c|c|c|c|c|c|}
\hline
\rowcolor{Gray}
\multicolumn{2}{|c|}{Models} & maF1 & miF1 & ebF1 & ACC & HA & Average \\
\hline
\rowcolor{Gray}
\multicolumn{8}{|c|}{Baselines} \\
\hline
\multicolumn{2}{|c|}{BR} &  0.349 & 0.480 & 	0.416 &  0.086 & \textbf{0.9957} & 0.465\\
\hline
\multicolumn{2}{|c|}{Seq2seq} & 0.345 & 0.448 &	0.421 & \textbf{0.140} &	0.9942 & 0.470 \\
\hline
\multicolumn{2}{|c|}{Seq2seq + SS} & 0.340 &	0.448 & 0.419 & 0.137 & 0.9943 & 0.468 \\
\hline
\multicolumn{2}{|c|}{Order-free RNN} &  0.310 & 0.438 & 0.410 & 0.096 & 0.9940 &	0.450 \\
\hline
\multicolumn{2}{|c|}{Order-free  RNN +  SS}  & 0.310 &	0.437 &	0.408 &	0.095 &	0.9947 &	0.449  \\
\hline
\rowcolor{Gray}
\multicolumn{8}{|c|}{Proposed methods} \\
\hline
\multicolumn{2}{|c|}{OCD} &  
\textcolor{blue}{0.353}  & 
0.465  &	
\textcolor{blue}{0.435}  & 0.117  &	0.9941  &	\textcolor{blue}{0.473} \\
\hline
\multirow{4}{*}{\shortstack{OCD\\+\\MTL}} & RNN dec. & 
\textcolor{blue}{0.359} & 	0.466 & 
\textcolor{blue}{0.438} & 0.115 & 	0.9940 & 
\textcolor{blue}{0.474} \\
\cline{2-8}
&BR dec. & 
\textcolor{blue}{0.353} &
\textcolor{blue}{0.485} &	0.420 &	0.075 &	0.9950 &	0.466 \\
\cline{2-8}
& Logistic rescore & 
\textcolor{blue}{\textbf{0.378}}  & 
\textcolor{blue}{0.487}	  & 
\textcolor{blue}{\textbf{0.456}}  &	0.096  &	0.9940  & 
\textcolor{blue}{0.482} \ \\
\cline{2-8}
& Logistic joint dec. & 
\textcolor{blue}{0.377} & 
\textcolor{blue}{\textbf{0.488}} &	
\textcolor{blue}{0.454} &	0.119 &	0.9945 &  \textcolor{blue}{\textbf{0.487}} \\
\hline
\end{tabular}}
\caption{Performance comparisons on Audio set.}
\vspace{-0.4cm}
\label{tab:audio_set}
\end{table}

\begin{figure}
  \includegraphics[width=\linewidth]{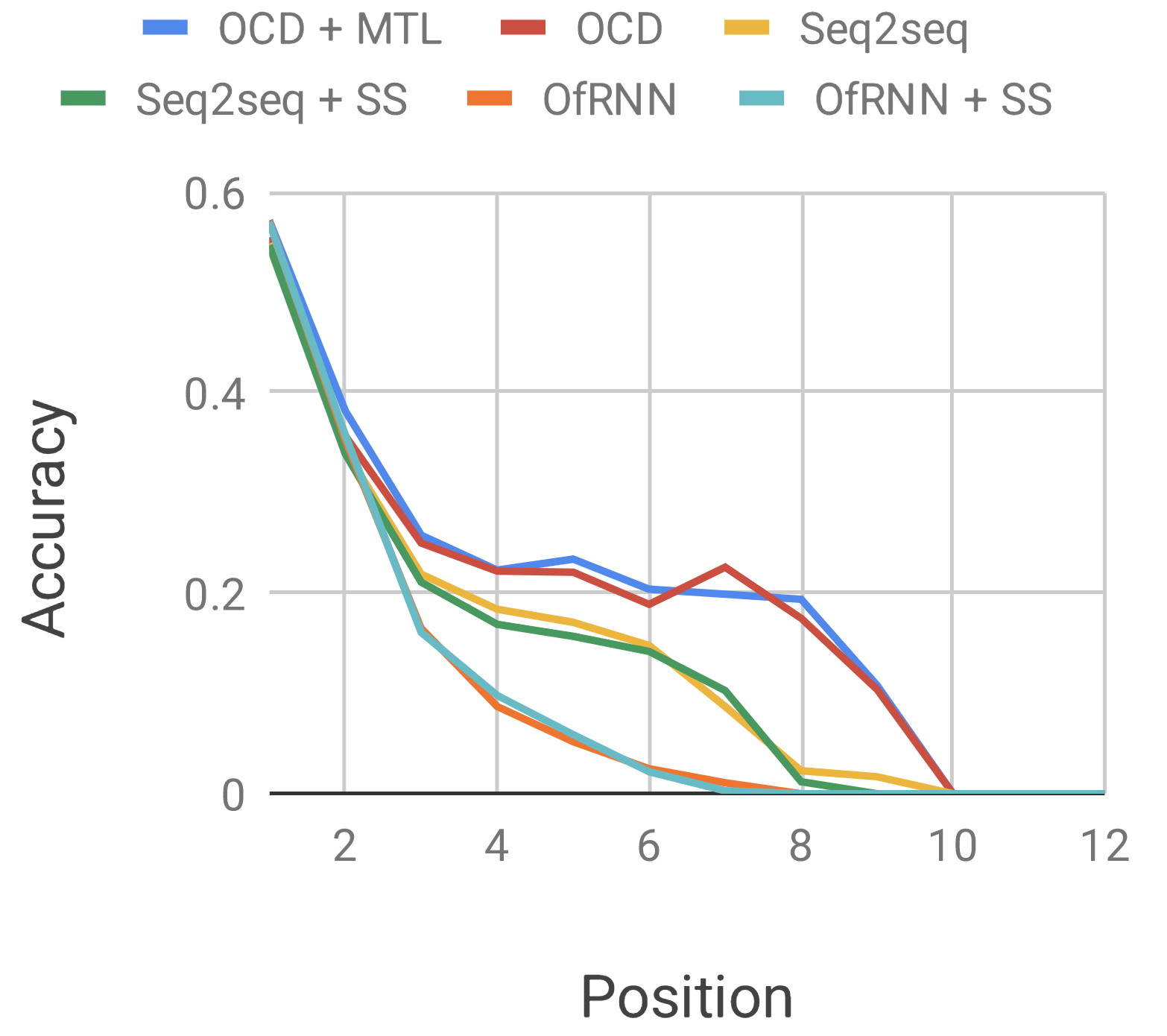}

  \caption{Position-wise accuracy of different models at each time step on Audio set. OCD+MTL was decoded by logistic joint decoding. Note that the length of the longest generated(reference) label sequence is 12.}
  \label{fig:error_propagation}
  \vspace{-0.4cm}
\end{figure}

For MTL, we report the performance for the decoding strategies from section \emph{Decoder Integration}. The first two decoding methods (rows (k),(l)) consider only the predictions of one decoder, while the other two (rows (m),(n)) combine the predictions using different decoding strategies. With MTL, we see the performance is improved across all the metrics except for ACC (row (j) v.s. row (k)). In addition, joint decoding methods (row(m),(n)) achieve the best performance, and  outperform previous works (row(a),(b),(c)). Interestingly, BR is also improved significantly with MTL (row (d) v.s. row(l)), the encoder of which may implicitly learn the label dependencies through the RNN decoder, which the original BR (row (d)) ignores.

Fig.~\ref{fig:training_curve} shows the validation ACC and miF1 curves for OCD, BR, and MTL with three decoding methods. BR performs the worst and converges the slowest. Nonetheless, with the help of MTL, BR converges much faster and better. Also, MTL helps to improve the performance of the OCD model.

\begin{table}
\centering
\resizebox{\columnwidth}{!}{
\begin{tabular}{|c|c|c|c|c|}
\hline
\rowcolor{Gray}
 & \multicolumn{2}{|c|}{Seen test set} &  \multicolumn{2}{|c|}{Unseen test set} \\
\hline
\rowcolor{Gray}
Models & miF1 & ebF1 &  miF1 & ebF1 \\
\hline
Seq2seq & 0.730 & 0.749 & 0.508 & 0.503 \\
\hline
Seq2seq + SS & 0.736 & 0.754 & 0.517 & 0.515 \\
\hline
Order-free RNN & 0.732 & 0.746 & 0.496 & 0.494 \\
\hline
Order-free RNN + SS & 0.724 & 0.740 &  0.520 & 0.517 \\
\hline
OCD (correct prefix) & 0.726 & 0.741 & 0.513	& 0.515 \\
\hline
OCD & \textbf{0.746} & \textbf{0.771} & \textbf{0.521} & \textbf{0.530} \\
\hline
\end{tabular}}
\caption{Performance comparison on resplited AAPD, whose test set contains 2000 samples whose label sets occur in the training set (Seen test set) and 2000 samples are not (Unseen test set). OCD (correct prefix) means we only sample correct labels in the training phase. }
\label{tab:resplited_aapd}
\vspace{-0.3cm}
\end{table}

\subsection{Experiments on Reuters-21758}

In comparison with AAPD, Reuters-21758 is a relatively small dataset. The average number of labels per sample is only 1.24 and over 80\% of the samples have only one label, making for a relative simple dataset.

Table~\ref{tab:reuters_performance} shows the results of the methods. These results demonstrate again the superiority of OCD and the performance gains afforded by the MTL framework. Since there are over 80\% of test samples only have one label in this corpus, to truly know the effect of  proposed approaches to multi-label classification, we also provide results only on test samples with more than one label in the Section \emph{Analysis of Reuters-21758} in Appendix.

\subsection{Experiments on Audio set}

\begin{table*}
\centering
\resizebox{\textwidth}{!}{
\begin{tabular}{|c|c|c|c|}
\hline
\rowcolor{Gray}
Models & Case 1 & Case 2 & Case 3 \\
\hline
Ground truth & \textbf{cs.it}, \textbf{math.it}, cs.ds & cs.lg, stat.ml, math.st, stat.th & \textbf{cs.it}, \textbf{math.it}, cs.ds, cs.dc \\
\hline
Seq2seq & \textbf{cs.it}, \textbf{math.it}, cs.ni & \textbf{cs.it}, \textbf{math.it}, math.st, stat.th & \textbf{cs.it}, \textbf{math.it}, cs.ni\\
\hline
Order-free RNN & \textbf{math.it}, \textbf{cs.it} & \textbf{math.it}, \textbf{cs.it}, stat.th, math.st & \textbf{math.it}, \textbf{cs.it}, cs.ni \\
\hline
OCD & \textbf{math.it}, \textbf{cs.it}, cs.ds   & stat.ml, stat.th, cs.lg, math.st & \textbf{cs.it}, \textbf{math.it} \\
\hline
OCD + MTL + joint dec. &  \textbf{cs.it}, \textbf{math.it}   &math.st, stat.th, stat.ml, cs.lg, \textbf{cs.it}, \textbf{math.it} & \textbf{math.it}, \textbf{cs.it} \\
\hline
\end{tabular}}
\caption{Examples of generated label sequences on AAPD from different models}
\vspace{-0.2cm}
\label{tab:example}
\end{table*}

\begin{table}
\centering
\resizebox{\columnwidth}{!}{
\begin{tabular}{|c|c|c|c|c|c|c|}
\hline
\rowcolor{Gray}
& Ref- &  & & Seq2seq &  & OfRNN \\
\rowcolor{Gray}
& erence & \multirow{-2}{*}{OCD} &\multirow{-2}{*}{Seq2seq} & + SS & \multirow{-2}{*}{OfRNN} & + SS \\
\hline
\rowcolor{Gray}
\multicolumn{7}{|c|}{AAPD} \\
\hline
$S_{test}$ & 392 & 302 & 214 & 293 & 251 & 259\\
\hline
$S_{test-train}$ & 43 & 30 & 1 & 3 & 1 & 4 \\
\hline
\rowcolor{Gray}
\multicolumn{7}{|c|}{Reuters-21758} \\
\hline
$S_{test}$ & 210 & 159 & 135 & 140 & 139 & 144  \\
\hline
$S_{test-train}$ & 94 & 37 &  15 & 16 & 23 & 26  \\
\hline
\rowcolor{Gray}
\multicolumn{7}{|c|}{Audio set} \\
\hline
$S_{test}$ & 6300 & 4787 & 1974 & 1781 & 1806 & 1789  \\
\hline
$S_{test-train}$ & 2445 & 1943 &  4 & 8  & 263 & 237  \\
\hline
\end{tabular}
}
\caption{Number of different generated (or reference) label combinations ($S_{test}$), and the number of generated (or reference) label combinations that do not occur in the training label sets ($S_{test-train}$) on three datasets.}
\label{tab:number_of_combinations}
\vspace{-0.3cm}
\end{table}

Table \ref{tab:audio_set} shows the performance of each model. In this experiment, all models have similar performance in HA. Surprisingly, BR is a competitive baseline model and performs especially well in miF1. Seq2seq achieves the best performance in terms of ACC, which is the same as the observation on AAPD. Overall,  OCD performs better than all the baseline and MTL indeed improves the performance. OCD outperforms other RNN decoder-based models in maF1, miF1 and ebF1 and performs worse than BR only in terms of miF1.

\subsection{Discussion}

\subsubsection{Propagation of errors}


Fig.~\ref{fig:error_propagation} shows position-wise accuracy of different models at each time step of RNN decoder on Audio set. 
The accuracy is calculated by checking whether or not model's generated labels are in reference label sets and then averaging the errors at each time step. 
If a generated label sequence is less than the corresponding target label sequence, the unpredicted part of the sequence is considered wrong. 

We can see that accuracy of all models decreases dramatically along x-axis. 
Because the labels are generated sequentially, the models would condition on wrong generated prefix label sequence in the test stage.
This problem may be amplified when the generated sequence is longer because of accumulation of errors.
Compared with the baseline models, OCD+MTL and OCD perform better after position 2, which demonstrates that they are more robust against error propagation, or exposure bias. Similar phenomenon can be observed in AAPD ( Fig. \ref{fig:error_propagation_aapd} in Appendix).

\subsubsection{Effectiveness of scheduled sampling and OCD}

To demonstrate the effectiveness of scheduled sampling and OCD when dealing with exposure bias, we evaluate the performance of models when tackling with unseen label combinations, where models encounter unseen situations and the problem of exposure bias may be more severe.

In this experiment, since there are only 43 samples with unseen label combinations in original test set of AAPD, we resplited the  AAPD dataset: 47840 samples in training set, 4000 samples for validation set and test set, respectively. Both validation set and test set contain 2000 samples whose label sets occur in the training set and 2000 samples are not.

Table \ref{tab:resplited_aapd} shows the results on resplited AAPD. OCD (correct prefix) means we only sample correct labels in the training phase, so this model has not encountered wrong prefix during training. Clearly, all models perform worse on unseen test set. We can see that SS improves the performance significantly on the unseen test set for both seq2seq and order free RNN.
Additionally, OCD with correct prefix, which suffers from the exposure bias, performs worse in both case than OCD. They all demonstrate that sampling wrong labels from predicted distribution helps models become more robust when encountering rare situation.

SS for MLC has a potential drawback. The input labels of RNN decoders obtained by sampling could be labels which do not conform to the predefined order. This may mislead the model. However, there is no label ordering in OCD, so this problem does not exist.

On both seen and unseen test set, OCD performs the best since OCD not only alleviates exposure bias but also does not need predefined order. Results of five metrics and another experiment about exposure bias on AAPD can be found in Appendix.

\subsubsection{Problem of overfitting}

Table~\ref{tab:number_of_combinations} shows number of different generated label combinations ($S_{test}$), and the number of generated label combinations that do not occur in the training label sets ($S_{test-train}$) on three datasets.
Seq2seq and OfRNN produce fewer kinds of label combinations on AAPD and Reuters-21758. 
As they tend to ``remember'' label combinations, the generated label sets are most alike, indicating a poor generalization ability to unseen label combinations. 
Because seq2seq is conservative and only generates label combinations it has seen in the training set, it achieves the highest ACC in Tables~\ref{tab:aapd_performance} and~\ref{tab:audio_set}.
For models with SS, they produce more kinds of label combinations, except for Audio set. 
OCD produces the most unseen label combinations on three datasets, since it encounters different label permutations during training. 

\subsubsection{Case study} 

Table~\ref{tab:example} shows examples of generated label sequences using different models on AAPD. Note labels \textit{cs.it} and \textit{math.it} in the three cases: Seq2seq produces label sequences only from frequent to rare, which is the same as the ground truth order, while order-free RNN learns the order implicitly. In contrast, OCD generates label sequences with flexible orders because it encounters different label permutations in the sampling process during training.

\section{Conclusion}
\label{sec:conclusion}
In this paper, we propose a new framework for multi-label classification based on optimal completion distillation and multitask learning. Extensive experimental results show that our method outperforms competitive baselines by a large margin. Furthermore, we systematically analyze exposure bias and the effectiveness of scheduled sampling.  

\bibliography{ref}
\bibliographystyle{aaai}

\clearpage

\section{Appendix}
\label{sec:appendix}
\subsection{Derivation of logistic joint decoding}
\label{sec:derivation}
In this appendix, we derive the equation for logistic joint decoding (Eq.~\ref{eq:br_joint_decoding}). We first reformulate Eq.~\ref{eq:p_br}.

\begin{equation}
	P_{br}(\mathcal{H}) = \prod_{l \in \mathcal{H}} \frac{p_{br}(y_{l}=1|\mathbf{x})}{p_{br}(y_{l}=0|\mathbf{x})} \times \prod_{l}p_{br}(y_{l}=0|\mathbf{x})  \\
\label{eq:reformulate_br}
\end{equation}

Since the second term of Eq.~\ref{eq:reformulate_br}  does not depend on $\mathcal{H}$, we can substitute it into Eq.~\ref{eq:p_joint}.

\begin{equation}
  \begin{split}
	& \hat{H} =  \arg\max_{\mathcal{H}}\{P_{path}(\mathcal{H}) \times P_{br}(\mathcal{H}) \} \\
    & = \arg\max_{\mathcal{H}} \{\prod_{i=1}^{i=T}p_{rnn}(l_i|\mathbf{x},l_1,...l_{i-1})\frac{p_{br}(y_{l_i}=1|\mathbf{x})}{p_{br}(y_{l_i}=0|\mathbf{x})}  \} \\
    \end{split}
\label{eq:br_joint_decoding_appendix}
\end{equation}

\subsection{Datasets and Preprocessing}
\label{sec:dataset}

\begin{table*}
\centering
\begin{tabular}{|c|c|c|c|c|c|c|}
\hline
\rowcolor{Gray} 
Dataset & $N_{training}$ & $N_{val}$ & $N_{test}$&  \#  labels & Words/sample & Labels/sample \\
\hline
Reuters-21758 & 6,993 & 776 & 3019 & 90 & 53.94 & 1.24 \\
\hline
AAPD & 53,840 & 1000 & 1000 & 54 & 163.42 & 2.41 \\
\hline
RCV1-V2 & 802,414 & 1000 & 1000 & 103 & 123.94 & 3.24 \\
\hline
Audio set & 2,063,949 & 0 & 20,371 & 527 & - & 1.98 \\
\hline
\end{tabular}
\caption{Summary of datasets. \# of training samples ($N_{training}$), \# of validation samples, ($N_{val}$), \# of test samples ($N_{test}$), \# of labels. Words/sample is the average number of labels per sample and labels/sample is the average number of labels per sample.}
\label{tab:dataset}
\end{table*}

We used three multi-label text classification datasets and one sound event classification dataset. The statistics of the four datasets are shown in Table~\ref{tab:dataset}. 

\begin{itemize}
\item \textbf{Reuters-21758}\footnote{http://www.daviddlewis.com/resources/testcollections/\\reuters21578/}: The Reuters-21758 dataset is a collection of around 10,000 documents that appeared on Reuters newswire in 1987 with 90 classes. 

\item \textbf{Arxiv Academic Paper Dataset (AAPD)}: This dataset is provided by \citet{yang2018sgm}, and consists of the abstracts and corresponding subjects of 55,840 academic computer science papers from arxiv. Each paper has 2.41 subjects on average.

\item \textbf{Reuters Corpus Volume I (RCV1-V2)}: The RCV1-V2 dataset~\cite{lewis2004rcv1} consists of a large number of manually categorized newswire stories (804,414) with 103 topics. 

\item \textbf{Audio Set}: 
The Audio set was proposed by Google \cite{gemmeke2017audio}, which consists of over 2 million 10-second audios covering 527 kinds of audio events, including music, speech, vehicles and animal sounds. Because Google only released bottleneck features provided by a pretrained Resnet-50, we used the features as inputs of models. The inputs are ten 128-dim bottleneck features for each audio clip.

\end{itemize}

When preparing Reuters-21758, we followed~\citet{nam2017maximizing} in randomly setting aside 10\% of the training instances as the validation set. For AAPD and RCV1-V2, we used the training/validation/test set split from \citet{yang2018sgm}.  For these three text datasets, we filtered out samples with more than 500 words, which removed about 0.5\% of the samples in each dataset. For Audio set, we follow \citet{yu2018multi} to use the original training/test set split from Google (no validation set).

\subsection{Experimental Settings}
\label{sec:exp_setting}

We implemented our experiments in Pytorch. Some hyperparameters of the model on four datasets are shown in Table~\ref{tab:hyp}. 

We used the Adam optimizer~\cite{kingma2014adam} with a learning rate of 0.0005. In addition, to avoid overfitting, we utilized dropout~\cite{srivastava2014dropout} and clipped the gradients to the maximum norm of 10. For OCD models, we set the softmax temperature $\tau$ to $10^{-8}$, which resulted in hard targets. 
For models with scheduled sampling, we set the teacher forcing ratio from 1.0 (start-of-training) to 0.7 (end-of-training). For MTL models, the weight $\lambda$ between OCD and logistic losses was 1. 

Different settings of three multi-label classification datasets and Audio set are in following.

\subsubsection{Multi-label text classification}
The BR decoder is a 3-layer DNN with 512 leaky-RELU units. The word embeddings were initialized randomly; their size was 512. 

During training, we trained the model for a fixed number of epochs and monitored its performance on the validation set every 1000 updates. Once the training was completed, we chose the model with the best micro-F1 score on the validation set and evaluated its performance on the test set.

During testing, we set the beam size to 6 during the beam search process for all the RNN decoders. For models with BR decoders, we chose the best threshold on the validation set to maximize the micro F1 score, and selected labels whose score was higher than the selected threshold~\cite{tu2018learning,quevedo2012multilabel}.

All the hyperparameters were tuned on the baseline model, seq2seq, until the performance is approximately equal to the performance of the same model reported in the previous works\cite{yang2018sgm,yang2018deep}. Then we applied the same model architecture and hyperparameters to all of the models for fair comparison.

\subsubsection{Multi-label sound event classification}

In this experiment, for the BR model, we used the architecture  provided by \citet{yu2018multi}
\footnote{https://github.com/qiuqiangkong/audioset\_classification} (2-A-1-A model). 
To the best of our knowledge, it achieved the best performance on Audio set. We reimplemented the model and achieved similar performance (Mean average precision  score \footnote{Because mean average precision measures ranking of confidence scores of each label, which RNN-based approaches can not generate since it only produces hard target sequences. Therefore, we did not utilize it as an evaluation metric.} 
of 0.349 without aggregating the output probabilities of models during training at each epoch).

\citet{yu2018multi} trained their model with binary logistic loss (Eq. \ref{eq:logistic_loss}). 
We decompose it into two parts: a BR decoder, which is the final fully-connected output layer with sigmoid activation function of their proposed model, and an encoder (the remaining part; without final output layer). 
For RNN-based model, we set the output of the encoder as the initial state of RNN decoder, which is comprised of 2 layers of LSTM with 512 hidden units. We used the technique of mini-batch balancing \cite{kong2018audio}.

Because we follow the setting of \cite{yu2018multi} and there is no validation set,  we trained models for 10 epoches and then test them on the test set. 
We set the beam size to 6 during the beam search process for all the RNN decoders. For models with BR decoders, we fix the threshold to $0.5$.

\begin{table}
\centering
\resizebox{\columnwidth}{!}{
\begin{tabular}{|c|c|c|c|c|}
\hline
\rowcolor{Gray} 
 & Vocab & LSTM & Batch &  \\
\rowcolor{Gray} 
\multirow{-2}{*}{Dataset} & size & layer & size & \multirow{-2}{*}{Dropout}  \\
\hline
Reuters & 22747 & (2,2) & 96 & 0.5 \\
\hline
AAPD & 30000 & (2,2) & 128 & 0.5 \\
\hline
RCV1-V2 & 50000 & (2,3) & 96 & 0.3 \\
\hline
Audio set & - & (-,2) & 500 & 0.5 \\
\hline
\end{tabular}}
\caption{Hyperparameters for datasets. LSTM layer (2,3) means the numbers of layers of the RNN encoder and decoder are 2 and 3, respectively. "-" means it does not exist.}
\label{tab:hyp}
\end{table}

\subsection{Evaluation Metrics}
\label{sec:metrics}
The five metrics can  be split in two parts.

\textbf{Example-based measures} are defined by comparing the target vector $\mathbf{y}^{*}$ to the prediction vector $\hat{\mathbf{y}}$. \textit{Subset accuracy} (ACC) is the most strict metric, indicating the percentage of samples in which all the labels are classified correctly. $\mathrm{ACC}(\mathbf{y}^{*},\hat{\mathbf{y}}) = \Bbb{I}[\mathbf{y}^{*}=\hat{\mathbf{y}}]$.
\textit{Hamming accuracy} (HA) is the 
ratio of the number of correctly predicted labels to the total number of labels. $\mathrm{HA}(\mathbf{y}^{*},\hat{\mathbf{y}})=\frac{1}{L}\sum_{i=1}^{i=L}\Bbb{I}[\mathbf{y}^{*}_{i}=\hat{\mathbf{y}}_{i}] $.
\textit{Example-based F1} (ebF1) defined by Eq.~\ref{eq:ebf1} measures the ratio of the number of correctly predicted labels to the total of the predicted and ground truth labels.

\begin{equation}
\ebF1(\mathbf{y}^{*},\hat{\mathbf{y}}) = \frac{2\sum^{L}_{i=1}\mathbf{y}^{*}_{i}\hat{\mathbf{y}}_{i} }{\sum^{L}_{i=1}\mathbf{y}^{*}_{i} +\sum^{L}_{i=1}\hat{\mathbf{y}}_{i}}
\label{eq:ebf1}
\end{equation}

\textbf{Label-based measures} treat each label $\mathbf{y}^{*}_i$ as a separate two-class prediction problem, and compute the number of true positives (tp), false positives (fp), and false negatives (fn) for each label over the dataset. 

\textit{Macro-averaged F1} (maF1) is the average of the F1 scores of each label (Eq.~\ref{eq:maf1}), and
\textit{Micro-averaged F1} (miF1) is calculated by summing tp, fp, and fn and then calculating the F1 score (Eq.~\ref{eq:mif1}).

High maF1 scores usually indicate high performance on rare labels, while high miF1 scores usually indicate high performance on frequent labels~\cite{nam2017maximizing}. 

\begin{equation}
maF1(\mathbf{y}^{*},\hat{\mathbf{y}}) = \sum^{L}_{i=1}\frac{2{tp}_{i}}{2{tp}_{i}+{fp}_{i}+{fn}_{i}}
\label{eq:maf1}
\end{equation}

\begin{equation}
miF1(\mathbf{y}^{*},\hat{\mathbf{y}}) = \frac{\sum^{L}_{i=1}2{tp}_{i}}{\sum^{L}_{i=1}2{tp}_{i}+{fp}_{i}+{fn}_{i}}
\label{eq:mif1}
\end{equation}

\subsection{Experiment on RCV1-V2}

\begin{table}
\centering
\resizebox{1.06\columnwidth}{!}{
\begin{tabular}{|c|c|c|c|c|c|c|c|}
\hline
\rowcolor{Gray}
\multicolumn{2}{|c|}{Models} & maF1 & miF1 & ebF1 & ACC & HA & Average \\
\hline
\rowcolor{Gray}
\multicolumn{8}{|c|}{Baselines} \\
\hline
\multicolumn{2}{|c|}{BR} & 0.671 & 0.868 & 0.881 & 0.642 & 0.9919 & 0.811 \\
\hline
\multicolumn{2}{|c|}{BR++} & 0.650 & 0.867 & 0.881 & 0.646 & 0.9919 & 0.807 \\
\hline
\multicolumn{2}{|c|}{Seq2seq} & 0.654 & 0.864 & 0.881 & \textbf{0.662} & 0.9916 & 0.811 \\
\hline
\multicolumn{2}{|c|}{Seq2seq + SS} & 0.653 & 0.860 & 0.878 &	0.658 &	0.9914 & 0.809 \\
\hline
\multicolumn{2}{|c|}{Order-free RNN} & 0.660 & 0.863 & 0.878 & 0.650 & 0.9917 & 0.809 \\
\hline
\multicolumn{2}{|c|}{Order-free  RNN +  SS}  & 0.637 & 0.862	& 0.876 & \textbf{0.662} & 0.9917 & 0.806 \\
\hline
\rowcolor{Gray}
\multicolumn{8}{|c|}{Proposed methods} \\
\hline
\multicolumn{2}{|c|}{OCD} & 0.668 & 0.866 & 
\textcolor{blue}{0.882} & 0.654 & 0.9918 &
\textcolor{blue}{0.812} \\
\hline
\multirow{4}{*}{\shortstack{OCD\\+\\MTL}} & RNN dec. & 
\textcolor{blue}{0.671} & 0.867 & 
\textcolor{blue}{0.882} & 0.651 & 0.9918 & 
\textcolor{blue}{0.813} \\
\cline{2-8}
&BR dec. & 0.663 & \textcolor{blue}{0.869} &	\textcolor{blue}{\textbf{0.885}} & 0.637 & \textcolor{blue}{\textbf{0.9920}} & \textcolor{blue}{0.813} \\
\cline{2-8}
& Logistic rescore & 
\textcolor{blue}{\textbf{0.676}} & 
\textcolor{blue}{0.869} & 
\textcolor{blue}{0.884} & 0.653 & 
\textcolor{blue}{0.9919} & 
\textcolor{blue}{0.815} \\
\cline{2-8}
& Logistic joint dec. & 
\textcolor{blue}{0.674} & 
\textcolor{blue}{\textbf{0.871}} & 
\textcolor{blue}{\textbf{0.885}} & 0.658 & \textcolor{blue}{\textbf{0.9920}} & 
\textcolor{blue}{\textbf{0.816}} \\

\hline
\end{tabular}}
\caption{Performance comparisons on RCV1-V2.}
\label{tab:rcv1_v2}
\end{table}

Table \ref{tab:rcv1_v2} shows the results. Compared to AAPD and Reuters-21578, RCV1-V2 consists of a larger number of documents. Moreover, the labels in this dataset have a hierarchical structure. If a leaf label belongs to one document, all labels from the root to the leaf label in the label tree also belong to the document. Hence, if we sort the labels from frequent to rare, parent labels precede child labels in the label tree.

In this dataset, OCD shows a smaller performance gain. This may be because the predefined order contains rich information about hierarchical structures which OCD does not utilize. However, datasets whose label have hierachical structures are not common.

\subsection{Detailed results of resplited AAPD}

\begin{table}[h]
\centering
\resizebox{1.06\columnwidth}{!}{
\begin{tabular}{|c|c|c|c|c|c|c|}
\hline
\rowcolor{Gray}
Models & maF1 & miF1 & ebF1 & ACC & HA & Average \\
\hline
\rowcolor{Gray}
\multicolumn{7}{|c|}{Seen test set} \\
\hline
Seq2seq & 0.530 & 0.730 & 0.749 & 0.453	&  0.9771 & 0.688 \\
\hline
Seq2seq + SS & 0.551 & 0.736 & 0.754 & 0.449 & 0.9774 & 0.693 \\
\hline
Order-free RNN & 0.545 & 0.732 & 0.746	 & \textbf{0.468} & 0.9777 & 0.694 \\
\hline
Order-free RNN + SS & 0.546 & 0.724 &  0.740 & 0.415 & 0.9764 & 0.680 \\
\hline
OCD(correct prefix) & 0.543 &	0.726 &	0.741 &	0.452 & 0.9770  & 0.688 \\
\hline
OCD & \textbf{0.571} & \textbf{0.746} & \textbf{0.771} & 0.443 & \textbf{0.9780} & \textbf{0.702} \\
\hline
\rowcolor{Gray}
\multicolumn{7}{|c|}{Unseen test set} \\
\hline
Seq2seq  & 0.402  & 0.503  & 0.508  & 0.002  & 0.9550  & 0.474 \\
\hline
Seq2seq + SS  & 0.418  & 0.515  & 0.517  & 0.009  & 0.9562  & 0.483 \\
\hline
Order free RNN  & 0.391  & 0.494  &	0.496  & 0.006  & 0.9560  & 0.469 \\
\hline
Order free RNN + SS  & 0.426 & 0.517 &	0.520  & 0.040  & 0.9557  & 0.492 \\
\hline
OCD(correct prefix) &	0.421 &	0.513 &	0.515 &	0.006 &	\textbf{0.9566}	& 0.482 \\
\hline
OCD & \textbf{0.446}  & \textbf{0.521}  & \textbf{0.530}  & \textbf{0.017}  & 0.9553  & \textbf{0.494} \\
\hline
\rowcolor{Gray}
\hline
\end{tabular}}
\caption{Detailed results on resplited AAPD (Table \ref{tab:resplited_aapd}).}
\label{tab:detailed_resplited_aapd}
\end{table}

Table \ref{tab:detailed_resplited_aapd} shows the detailed results of Table \ref{tab:resplited_aapd}. Interestingly, all models have difficulties predicting all the correct labels of unseen label combinations. Hence, the subset accuracy is extremely low on unseen test set. 

\subsection{Impact of exposure bias on AAPD}

\begin{figure}
  \includegraphics[width=\linewidth]{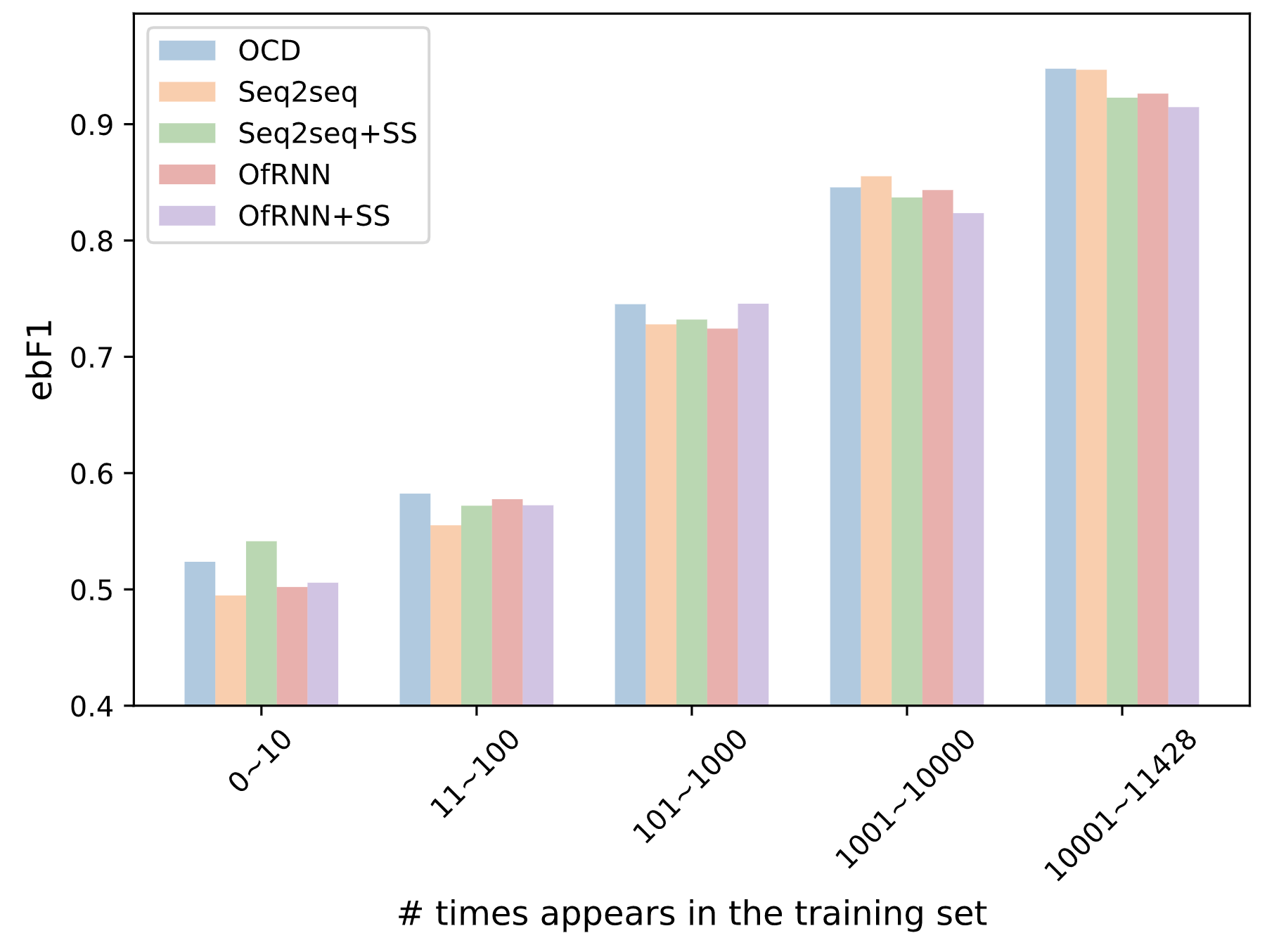}

  \caption{Example-based F1 score of test samples versus the number of times the label combination appears in the training set on AAPD. ``OfRNN'' denotes order-free RNN. }
  \vspace{-0.2cm}
\label{fig:ebf1_vs_freq}
\end{figure}

Fig.~\ref{fig:ebf1_vs_freq} shows the example-based F1 score of test samples of different models versus the number of times that the label combination appears in the training set on AAPD. 
Clearly, the more times the model sees the label combinations, the higher the performance. 
An interesting observation is that scheduled sampling (SS) helps Seq2seq and OfRNN with rare label combinations but not with frequent ones. 
This may be because with models that perform poorly when presented with rare situations, exposure bias may become more severe and models are more likely to make wrong predictions. Hence, SS is more helpful with rare examples.

SS performs worse when presented with frequent label combinations. 
For models with SS, labels obtained by sampling may be  labels which do not conform to the predefined order, which may mislead the model. 
In contrast, OCD performs well consistently. Since in OCD the loss function depends on the input prefixes, and we never supply the ground truth sequence, the model explores more states at the training stage. Hence, it is more robust under all situations. 

\subsection{Analysis of Reuters-21758}
\begin{table}[h]
\centering
\resizebox{1.06\columnwidth}{!}{
\begin{tabular}{|c|c|c|c|c|c|c|c|}
\hline
\rowcolor{Gray}
\multicolumn{2}{|c|}{Models} & maF1 & miF1 & ebF1 & ACC & HA & Average \\
\hline
\rowcolor{Gray}
\multicolumn{8}{|c|}{Baselines} \\
\hline
\multicolumn{2}{|c|}{BR} & 0.315 &	0.706 & 0.712 & 0.365 & 0.9850 &	0.617 \\
\hline
\multicolumn{2}{|c|}{Seq2seq} & 0.316 & 0.712 & 0.718 & 0.405 & 0.9855 & 0.627 \\
\hline
\multicolumn{2}{|c|}{Seq2seq+SS} & 0.325 & 0.718 & 0.722 & 0.380 & 0.9859 &	0.626 \\
\hline
\multicolumn{2}{|c|}{Order-free RNN} & 0.331 & 0.730 &	0.735 & 0.425 & 0.9862 & 0.641 \\
\hline
\multicolumn{2}{|c|}{Order-free  RNN +  SS}	& 0.324 & 0.699	 & 0.711 & 0.400 &	0.9849	 & 0.624 \\
\hline
\rowcolor{Gray}
\multicolumn{8}{|c|}{Proposed methods} \\
\hline
\multicolumn{2}{|c|}{OCD} & 0.319 & 
\textcolor{blue}{0.734} & 
\textcolor{blue}{0.741} & 0.415 & 
\textcolor{blue}{0.9864} & 
\textcolor{blue}{0.639} \\
\hline
\multirow{4}{*}{\shortstack{OCD\\+\\MTL}} & RNN dec. & 
\textcolor{blue}{0.335} &	
\textcolor{blue}{0.745}	& 
\textcolor{blue}{0.749} & 
\textcolor{blue}{\textbf{0.440}} & 
\textcolor{blue}{\textbf{0.9870}}	& 
\textcolor{blue}{0.651} \\
\cline{2-8}
& BR dec. & 0.322 &	
\textcolor{blue}{0.739} &	
\textcolor{blue}{0.737} &
\textcolor{blue}{0.430} & 
\textcolor{blue}{0.9869}	 & 
\textcolor{blue}{0.643} \\
\cline{2-8}
& Logistic rescore & 
\textcolor{blue}{0.337} & 
\textcolor{blue}{\textbf{0.750}}	 & \textcolor{blue}{\textbf{0.752}}  &	
\textcolor{blue}{0.435 } & 
\textcolor{blue}{0.9869}  &	
\textcolor{blue}{\textbf{0.652}} \\
\cline{2-8}
& Logistic joint dec. & 
\textcolor{blue}{\textbf{0.342}} &	
\textcolor{blue}{0.743} &
\textcolor{blue}{0.746} &
\textcolor{blue}{0.435} & 
\textcolor{blue}{\textbf{0.9870}}	& 
\textcolor{blue}{0.651} \\
\hline
\rowcolor{Gray}
\hline
\end{tabular}}
\caption{Performance comparisons on Reuters-21578 with more than one label.}
\label{tab:reuters_more_than_one}
\end{table}

Table \ref{tab:reuters_more_than_one} shows the results on the test set of Reuters-21758 with more than one label. The smaller test set has 405 samples. Comparing to Table \ref{tab:reuters_performance}, all models perform worse on this smaller test set since samples with only one label are taken out. However, the performance gap between baseline models and proposed methods are larger, which strengthen the superiority of OCD and MTL.

\subsection{Average ranking of models}

\begin{figure}[h]
  \includegraphics[width=\linewidth]{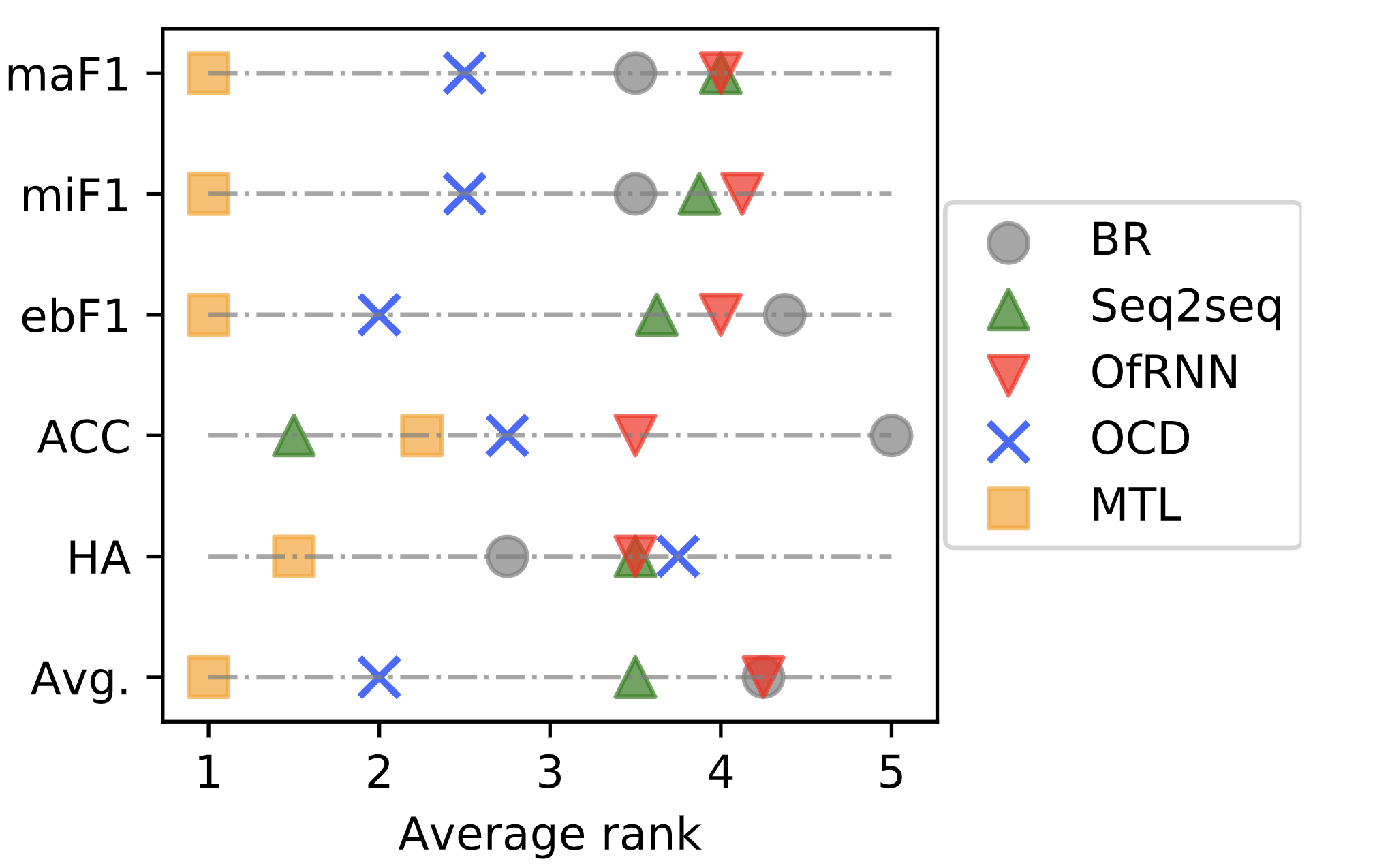}

  \centering
  \caption{Average ranks of different methods on four different datasets. The smaller the rank value, the better the performance. The MTL results are decoded by logistic joint decoding; ``OfRNN'' denotes order-free RNN.}
\label{fig:average_rank}
\vspace{-0.2cm}
\end{figure}

Fig.~\ref{fig:average_rank} shows the average ranks of four datasets using different metrics. Note that all models achieve similar performance on HA on these datasets.
Clearly, MTL performs the best, followed by OCD. Note that Seq2seq achieves the best performance in terms of ACC, but it performs worse in terms of other metrics. 

\subsection{Position-wise accuracy on AAPD}

\begin{figure}[h]
  \includegraphics[width=\linewidth]{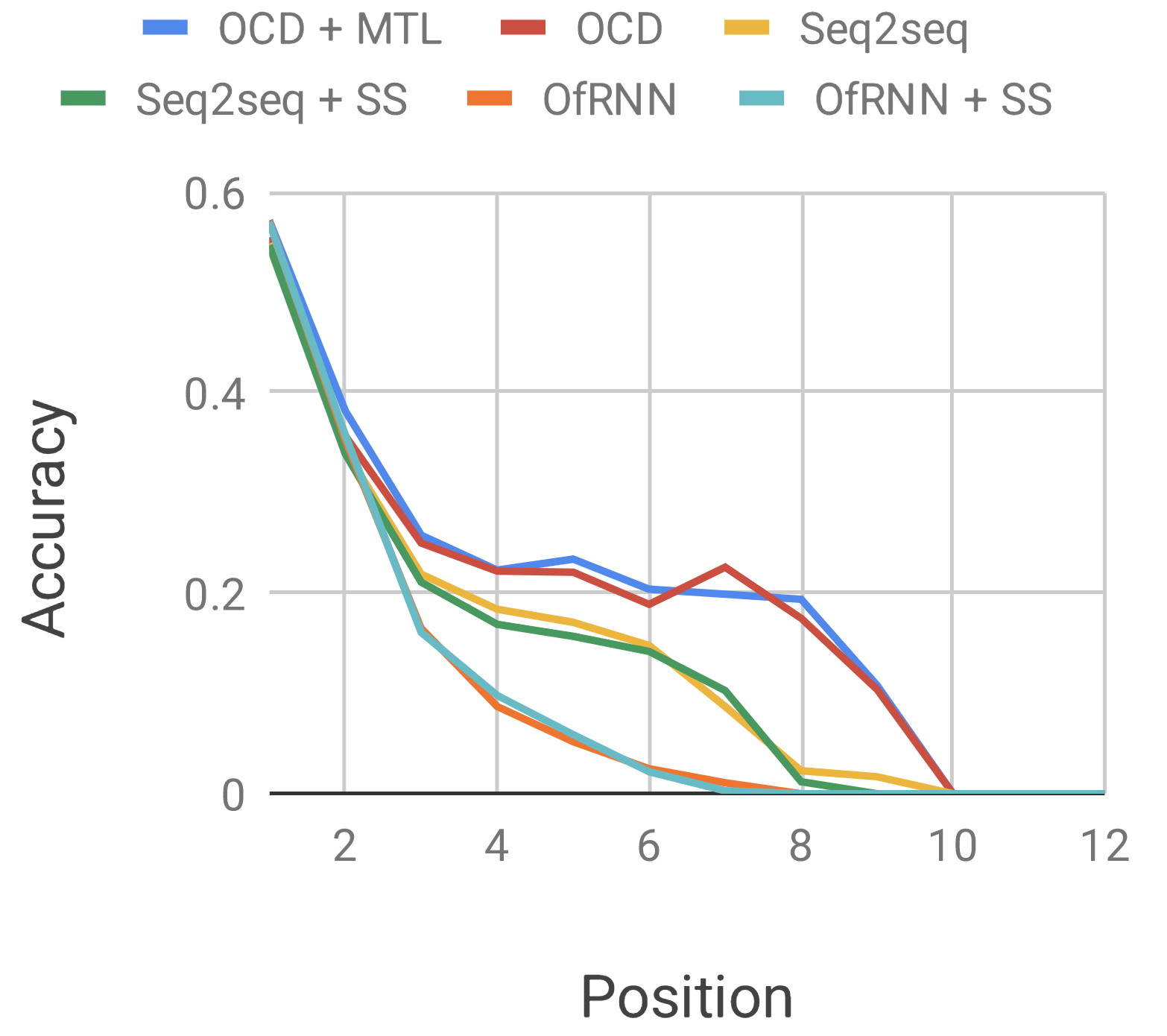}

  \caption{Position-wise accuracy of different models at each time step on AAPD. OCD+MTL was decoded by logisttic joint decoding. Note that the length of the longest generated(reference) label sequence is 6.}
  \label{fig:error_propagation_aapd}
\end{figure}

\end{document}